\documentclass[lettersize,journal]{IEEEtran}

\usepackage{CJKutf8}
\usepackage{times}
\usepackage{soul}
\usepackage{url}
\usepackage[hidelinks]{hyperref}
\usepackage[utf8]{inputenc}
\usepackage[small]{caption}
\usepackage{graphicx}
\usepackage{amsmath}
\usepackage{amsthm,bm}
\usepackage{algorithm,algpseudocode}
\usepackage[switch]{lineno}
\usepackage{comment}
\usepackage{amsfonts}
\usepackage{xcolor}
\usepackage{multirow, multicol}
\usepackage{colortbl,adjustbox,booktabs}
\usepackage{arydshln}
\usepackage{tikz}
\usepackage{subcaption}

\newcommand{\set}[1]{\ensuremath{\mathcal #1}}

\newcommand{\bfx}{{\mathbf x}}
\newcommand{\bfz}{{\mathbf z}}

\newcommand{\bmeps}{\bm{\epsilon}}

\newcommand{\bmtheta}{\bm{\theta}}

\newcommand\DPQ{\texttt{PQCAD-DM}}

\newcommand{\hr}[1]{[\textcolor{red}{HR: #1}]}

\usepackage[utf8]{inputenc}
    
\begin{document}

\title{{\DPQ}: Progressive Quantization and Calibration-Assisted Distillation for \\
Extremely Efficient Diffusion Model}

\author{Beomseok Ko,~\IEEEmembership{Student Member,~IEEE,} 
and Hyeryung Jang,~\IEEEmembership{Member,~IEEE}
\thanks{
This research was supported by the MSIT(Ministry of Science and ICT), Korea, under the ITRC(Information Technology Research Center) support program(IITP-2025-2020-0-01789), and the Artificial Intelligence Convergence Innovation Human Resources Development(IITP-2025-RS-2023-00254592) supervised by the IITP(Institute for Information \& Communications Technology Planning \& Evaluation). \textit{The first two authors contributed equally to this work.}

Beomseok Ko and Hyeryung Jang are with the Department of Computer Science \& Artificial Intelligence, Dongguk University, Seoul, South Korea (Corresponding Author: Hyeryung Jang) (emails: \{roy7001, hyeryung.jang\}@dgu.ac.kr)

}
}


\begin{CJK}{UTF8}{}
\CJKfamily{mj}

\maketitle

\begin{abstract}
Diffusion models excel in image generation but are computational and resource-intensive due to their reliance on iterative Markov chain processes, leading to error accumulation and limiting the effectiveness of naive compression techniques. 
In this paper, we propose {\DPQ}, a novel hybrid compression framework combining Progressive Quantization (PQ) and Calibration-Assisted Distillation (CAD) to address these challenges. 
PQ employs a two-stage quantization with adaptive bit-width transitions guided by a momentum-based mechanism, reducing excessive weight perturbations in low-precision. 
CAD leverages full-precision calibration datasets during distillation, enabling the student to match full-precision performance even with a quantized teacher. 
As a result, {\DPQ} achieves a balance between computational efficiency and generative quality, halving inference time while maintaining competitive performance. 
Extensive experiments validate {\DPQ}'s superior generative capabilities and efficiency across diverse datasets, outperforming fixed-bit quantization methods. 

\end{abstract}

\begin{IEEEkeywords}
Diffusion Models, Quantization, Distillation, Model Compression
\end{IEEEkeywords}

\section{Introduction} \label{sec:intro}

\IEEEPARstart{D}{iffusion} models (DMs)~\cite{diffusion} have emerged as a foundational technology for generative tasks, producing high-quality images with a balance of diversity and fidelity. 
They address issues such as mode collapse in GANs~\cite{gan,diffusionmodelsbeatgans} and blur artifacts in VAEs~\cite{vae}, making them highly effective for tasks ranging from image inpainting~\cite{inpainting,inpainting2} to graph generation~\cite{grap_gen} and molecular modeling~\cite{molecular}. 
However, the iterative generation process of DMs requires a large number of sampling steps, resulting in significant computational demands and slower inference speeds compared to other generative models. For instance, training state-of-the-art DMs can require hundreds of GPU days (e.g., $150$-$1000$ days on V100 in \cite{diffusionmodelsbeatgans}), making deployment challenging. This limits their practicality in resource-constrained environments, such as edge devices or real-time applications.

\begin{figure*}[t]
  \centering
  \includegraphics[width=0.95\textwidth]{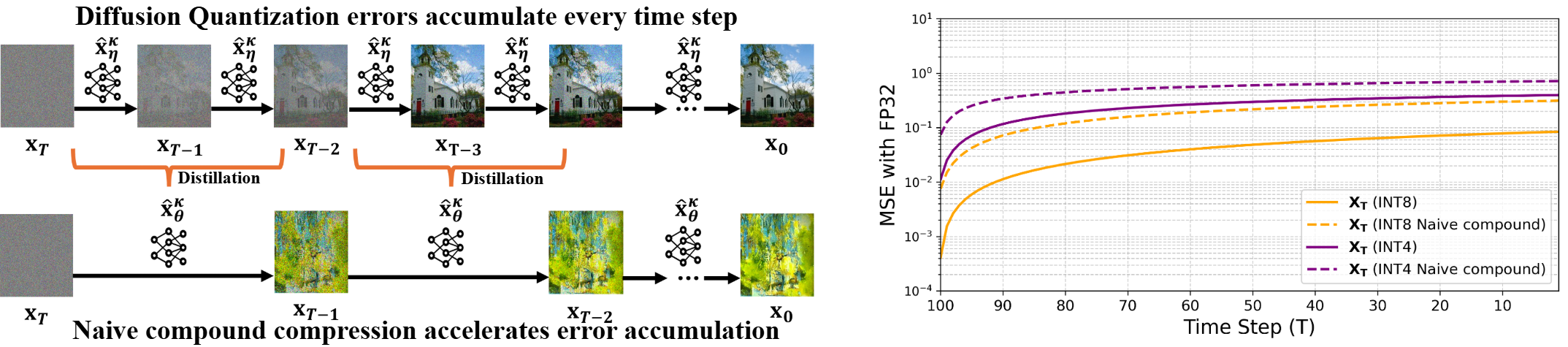}
  \caption{
  (Left) Compounding quantization and distillation in DMs introduces cumulative errors at each time step, significantly impacting generative performance. 
  (Right) Mean squared error (MSE) comparisons between the full-precision DDIM (100-time step ImageNet-trained) and both PTQ4DM quantized and naive compounded models highlight the need to mitigate error propagation during compression.
  }
  \label{fig:error}
\end{figure*}

Recent efforts have explored quantization and distillation to alleviate these computational challenges in two directions: reducing model size and accelerating sampling time. 
Quantization reduces the bit-width of weights and activations, lowering memory and computational costs, with Post-Training Quantization (PTQ) being particularly useful in diffusion models as it compresses models without requiring retraining~\cite{ptq4dm,q-diff}. 
At the same time, distillation transfers knowledge from a large teacher model to a smaller student model, accelerating sampling speed~\cite{progressivedistill,progressivedistill2}. 
Despite these advancements, combining these techniques effectively for diffusion models remains underexplored. 
Quantization can lead to cumulative errors, especially in iterative processes in DMs, and the sequence of applying distillation and quantization critically impacts performance if not carefully managed, as illustrated in Fig.~\ref{fig:error}.

In this paper, we propose \textbf{\DPQ}, a hybrid compression framework that combines Progressive Quantization and Calibration-Assisted Distillation to reduce the computational complexity of diffusion models. 
The core features of {\DPQ} include: {\em (i)} Progressive Quantization (PQ) that employs variable bit-widths to precisely control excessive weight perturbations at low bit precision, addressing the temporal variability in DMs with time-aware calibration datasets; and {\em (ii)} Calibration-Assisted Distillation (CAD) that mitigates inaccuracies inherited from the quantized teacher model during distillation by using a full-precision calibration dataset, ensuring minimal degradation in student model performance. 
This unified approach enhances quantization accuracy, prevents performance degradation during subsequent distillation, and strikes a balance between model quality and computational efficiency, achieving both high generative quality and improved inference speed.

Our contributions are summarized as follows:
\begin{itemize}
    \item We propose \textbf{\DPQ}, a hybrid compression framework that integrates progressive quantization and calibration-assisted distillation. This approach reduces model size and sampling steps, enabling faster inference while preserving the performance and generative quality of diffusion models.


    \item We introduce a progressive quantization strategy to minimize weight disturbances at low precision and incorporate full-precision calibration during distillation, effectively balancing the computational efficiency and generative accuracy of the final compressed DMs. 
    

    \item Extensive experiments on CIFAR-10, ImageNet, CelebA-HQ, and LSUN-Bedrooms/Churches datasets demonstrate that {\DPQ} achieves competitive image quality (regarding FID, sFID, and IS) while significantly reducing computational overhead, halving inference time and improving FID by 0.17 on average compared to quantization baseline methods.
    

\end{itemize}

\section{Related Works} \label{sec:pre}

\noindent \textbf{Efficient Diffusion Models.} 
Various approaches have been studied to improve the efficiency of diffusion models (DMs), including optimized sampling~\cite{kong2021fast,ddpm,ddim,iddim,iddpm}, pruning~\cite{diffusionpruning,ldm_pruning}, parameter quantization~\cite{ptq4dm,q-diff}, and knowledge distillation~\cite{progressivedistill,knowledgedistill}. 
Sampling methods like DDIM reduce generation time at the cost of resources and time for training, while pruning and post-training quantization (PTQ) methods reduce the model size and memory consumption without retraining. 
Distillation accelerates DM sampling by transferring knowledge from a teacher model (with long sampling steps) to a student model (with shorter steps). 
Despite their individual strengths, these approaches often focus on a single aspect of compression, such as sampling or quantization, and fail to address the cumulative errors that arise when combining compression methods.



\noindent \textbf{Progressive Quantization.} 
Low-precision quantization introduces significant challenges during training due to quantization noise, prompting the development of progressive quantization techniques. These methods gradually reduce precision (or bit-widths), starting with high precision and transitioning to lower bit-widths~\cite{towardsquant,zhuang2021effective,qu2020adaptive,kim2022ctmq,jung2019learning,bit-shrinking}. 
For instance, strategies like bit-shrinking~\cite{bit-shrinking} adaptively transition bit-widths, enabling simultaneous quantization of weights and activations with minimum errors. 
While effective in traditional networks, their application to DMs is limited due to the temporal variability of activations inherent in the diffusion process, requiring specialized strategies, such as time-aware calibration, to effectively manage the unique challenges posed by DM quantization.

\vspace{-0.1cm}
\section{Preliminary} \label{sec:prelim}

\noindent \textbf{Diffusion Models.} 
Diffusion models (DMs)~\cite{ddpm,variational} generate data (e.g., images) through a forward and reverse process (see Fig.~\ref{fig:error}). 
In the forward process, Gaussian noise $\bmeps \sim \mathcal{N}({\bm 0}, \mathbf{I})$ is gradually added to input data $\bfx \sim p_\text{data}(\bfx)$ over $T$ time steps, transforming it into a noisy version $\{\bfx_0 = \bfx, \bfx_1, \ldots, \bfx_T\}$. 
The reverse process removes the noise to reconstruct the original data. Given noise $\bmeps$ and noise schedule factors $\{\alpha_t, \sigma_t\}_{t=0}^T$, the DM $\hat{\bfx}_\eta$ (with trainable parameters $\eta$) takes $\bfz_t = \alpha_t \bfx + \sigma_t \bmeps$ as input and is trained by minimizing the weighted mean squared error between the original data $\bfx$ and the denoised output $\hat{\bfx}_\eta(\bfz_t)$:
\begin{align} \label{eq:loss_DM}
    \set{L}_\text{DM} = \mathbb{E}_{t, \bfx \sim p_{\text{data}}(\bfx),\bfz_t \sim q(\bfz_t | \bfx)} \Big[ w(\lambda_t) \| \bfx - \hat{\bfx}_{\eta}(\bfz_t) \|_2^2 \Big],
\end{align}
where $t \sim \text{U}[0,1]$ is (normalized) time-step; $q(\bfz_t | \bfx) = \mathcal{N}(\bfz_t; \alpha_t \bfx, \sigma_t^2 \mathbf{I})$; $\lambda_t = \log(\alpha_t^2/\sigma_t^2)$ denotes the signal-to-noise ratio (SNR); and $w(\cdot)$ is a pre-defined weight function. 
After training, the reverse process is efficiently handled by the discrete-time DDIM~\cite{ddim} sampler, which updates $\bfz_s$ using the model's predictions at each step $s \in [0,1]$. Starting from $\bfz_t \sim \mathcal{N}({\bm 0}, \mathbf{I})$, the updates are given by:
\begin{align} \label{eq:DDIM}
    \bfz_s = \alpha_s \hat{\bfx}_\eta(\bfz_t) + \sigma_s \big( \frac{\bfz_t - \alpha_s \hat{\bfx}_\eta(\bfz_t)}{\sigma_t} \big), ~\text{for}~ s = t - 1/T.
\end{align}

\begin{figure*}[t!]
  \centering{\includegraphics[width=\textwidth]{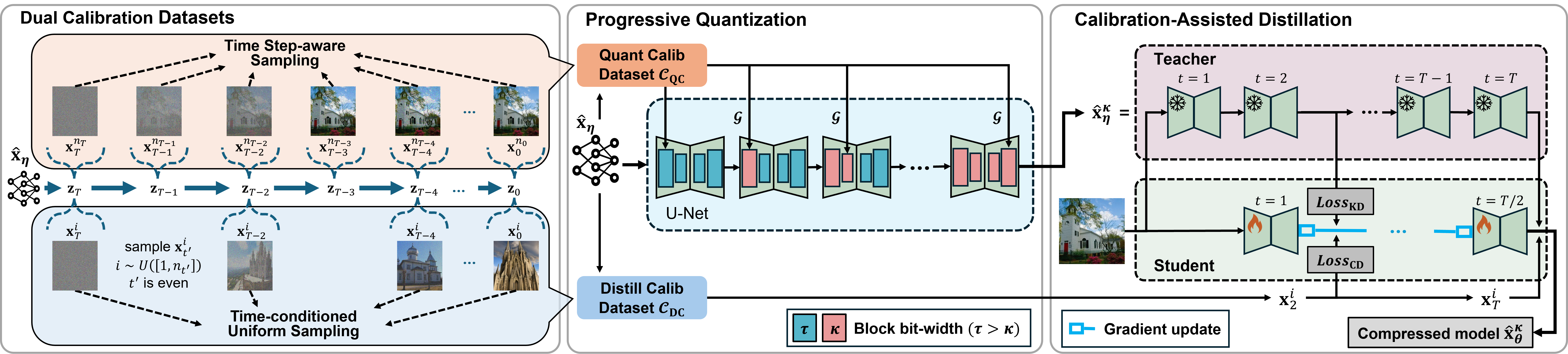}}
  \caption{
  Overview of {\DPQ} framework. (Left) Dual calibration datasets, (Center) Progressive Quantization (PQ), and (Right) Calibration-Assisted Distillation (CAD) optimize compression and generative quality.
}
  \label{fig:overall}
\end{figure*}

\noindent \textbf{Distillation of Diffusion.} 
Distillation techniques for diffusion models~\cite{progressivedistill,progressivedistill2} aim to reduce the number of sampling steps required in the reverse process, accelerating inference. 
The student model $\hat{\bfx}_\theta$ (with trainable parameters $\theta$) is trained to align one of its steps with two steps performed by the teacher model $\hat{\bfx}_\eta$, effectively halving the required sampling steps from $T$ to $T/2$. 
The teacher performs two DDIM sampling steps, first from $t$ to $t'=t-0.5/T$, and then from $t'$ to $t'' = t-1/T$. Instead of targeting the original data $\bfx$ as in \eqref{eq:DDIM}, the student learns to match the teacher's outcome of these two steps using a single step. The intermediate updates for the teacher are given by: 
\begin{align*}
    \bfz_{t'} &= \alpha_{t'} \hat{\bfx}_\eta(\bfz_t) + \frac{\sigma_{t'}}{\sigma_t} \big( \bfz_t - \alpha_t \hat{\bfx}_\eta(\bfz_t) \big), \cr 
    \bfz_{t''} &= \alpha_{t''} \hat{\bfx}_\eta(\bfz_{t'}) + \frac{\sigma_{t''}}{\sigma_{t'}} \big( \bfz_{t'} - \alpha_{t'} \hat{\bfx}_\eta(\bfz_{t'}) \big).
\end{align*}
The target $\tilde{\bfx}$ for the student is then computed as $\tilde{\bfx} = \frac{\bfz_{t''} - (\sigma_{t''}/\sigma_t) \bfz_t }{ \alpha_{t''} - (\sigma_{t''}/\sigma_t) \alpha_t }$. 
Using this target, the objective for student model $\hat{\bfx}_\theta$ is modified to: 
\begin{align} \label{eq:distill-DM}
    \set{L}_\text{Dist} = \mathbb{E}_{t, \bfx \sim p_{\text{data}}(\bfx),\bfz_t \sim q(\bfz_t | \bfx)} \Big[ w(\lambda_t) \| \tilde{\bfx} - \hat{\bfx}_{\theta}(\bfz_t) \|_2^2 \Big].
\end{align}

\begin{figure}[h!]
  \centering{\includegraphics[width=\linewidth]{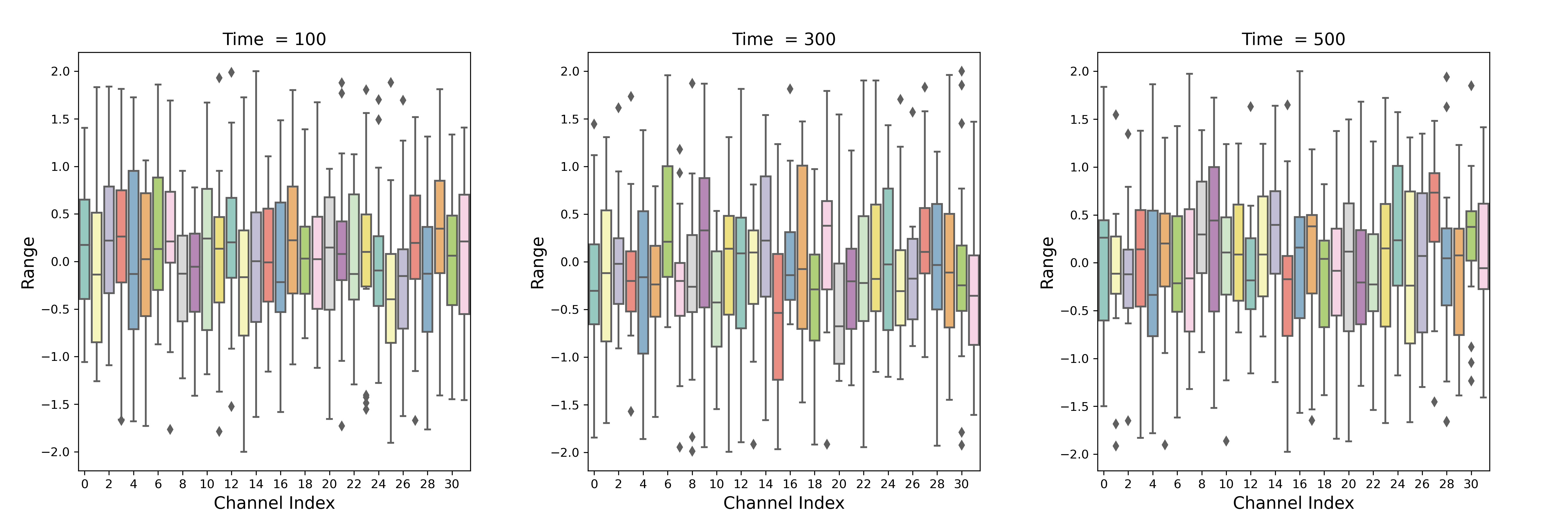}}
  \caption{Activation distribution over time step of diffusion. Figure shows the outputs of the first attention block of a DDIM model pre-trained on the CIFAR-10 dataset, presented using boxplots for each time step. A boxplot represents the median, max, and min values.
}
  \label{fig:image4}
\end{figure}

\noindent \textbf{Post-Training Quantization (PTQ) of Diffusion.} 
In diffusion models (DMs), the activation distribution of the noise estimation network varies significantly across time steps, as each step takes the output of the previous one as input. This temporal dependency causes the same network to exhibit drastically different activation patterns over time (Fig.~\ref{fig:image4}), particularly in deeper layers. 
As a result, PTQ in diffusion models (DMs) faces significant challenges due to activation variability across time steps.
Recent methods~\cite{ptq4dm,q-diff,tdq-diff,ptqd,q-dm,tfmq} address this issue by constructing {\em time-step-specific} calibration datasets that capture activation and weight distributions at each step, ensuring stable performance in quantized models. 
Q-Diff~\cite{q-diff} uses residual bottleneck and transformer blocks for reconstruction, calibrating weight quantizers and employing layer-wise reconstruction and split quantization to preserve UNet performance. 
PTQ4DM~\cite{ptq4dm} enhances this with block-wise reconstruction and introduces QDrop, a mechanism that randomly skips activation quantization to exploit model flatness. 
TFMQ-DM~\cite{tfmq} refines Q-Diff by addressing calibration overfitting and temporal feature variability with a temporal-aware reconstruction process. 
All methods maintain generative quality by prescaling weights during quantization and applying activation quantization to handle variability.

\noindent \textbf{Low-bit Post-Training Quantization.} 
Reconstruction-based quantization methods~\cite{adaptive_round,brecq} learn to adjust weights by rounding up or down, modeling the rounding operation as weight perturbation ($\hat{\mathbf{\eta}} = \mathbf{\eta} + \Delta \mathbf{\eta}$). 
For a pre-trained DM with weights $\eta$, these methods analyze the quantization objective, with $\hat{\eta}$ being quantized weights, using a Taylor expansion to capture the interactions between weights. The objective is expressed as:
\begin{equation} 
\min_{\hat{\mathbf{\eta}}} \mathbb{E} \Big[ \hat{\bfx}_{\hat{\eta}}(\bfz_t) - \hat{\bfx}_{\eta}(\bfz_t)\Big] \approx \mathbb{E} \Big[ \frac{1}{2} \Delta \mathbf{\eta}^\top \mathbf{H}^{\mathbf{\eta}} \Delta \mathbf{\eta} \Big],
\end{equation}
where \(\mathbf{H}^{\mathbf{\eta}} = \mathbb{E}\nabla_{\mathbf{\eta}}^2 ~\hat{\bfx}_{\eta}(\bfz_t)\) represents the expected Hessian of the output with respect to the weights $\eta$. This reformulation shows that the quantization objective depends on the change in weights, scaled by the output Hessian. The objective can further be approximated at the block or layer level:
\begin{equation} \label{eq:block_recon}
\min_{\hat{\mathbf{\eta}}} \mathbb{E} \Big[ \Delta \mathbf{\eta}^\top \mathbf{H}^{\mathbf{\eta}} \Delta \mathbf{\eta} \Big] \approx \mathbb{E} \Big[ \Delta \mathbf{a}^\top \mathbf{H}^{\mathbf{a}} \Delta \mathbf{a} \Big],
\end{equation}
where \(\mathbf{a}\) represents the output of a layer or block. By refining weights to minimize the error between quantized and full-precision outputs at the block level, these methods better address inter-layer dependencies and generalization compared to fully layer-wise corrections. This block-wise reconstruction improves overall performance, effectively balancing quantization noise and accuracy.

\section{METHOD} \label{sec:method}





\subsection{{\DPQ}: Overall Framework}

{\DPQ} is a hybrid compression framework designed to reduce the computational overhead of diffusion models in terms of time, energy, and memory, without significant performance degradation. It integrates two key components: 
\begin{enumerate}
    \item {\bf Progressive Quantization (PQ)}: PQ gradually reduces model bit-widths in stages to minimize quantization-induced errors. A momentum-based bit transition mechanism is employed for precise control of weight and activation quantization. 

    \item {\bf Calibration-Assisted Distillation (CAD)}: CAD mitigates inaccuracies inherited from quantized teacher models by utilizing a high-quality calibration dataset, allowing the student model to maintain efficiency and accuracy in the inference process. 
\end{enumerate}

As illustrated in Fig.~\ref{fig:overall}, {\DPQ} sequentially applies these techniques, reducing both model size and sampling steps. 
In the PQ phase, the pre-trained diffusion model $\hat{\bfx}_\eta$ is first quantized to $\tau$-bit precision, producing $\hat{\bfx}_\eta^\tau$. A second stage of quantization further reduces the precision of weights and activations to $\kappa$-bits, yielding $\hat{\bfx}_\eta^\kappa$. 
In the CAD phase, the $\kappa$-bit teacher model with $T$ sampling steps is distilled to produce the student model $\hat{\bfx}_\theta^\kappa$, which requires only $T/2$ inference steps. 
This sequential approach ensures that errors introduced during quantization are mitigated before distillation, achieving a trade-off between computational efficiency and generative quality.

\subsection{Dual Calibration Datasets} \label{sec:dataset}

As shown in Fig.~\ref{fig:overall} (left), {\DPQ} employs two distinct calibration datasets to ensure accurate quantization and effective distillation.

\noindent \textbf{Quantization Calibration Dataset.} 
In DMs, activation distributions vary significantly across time steps, necessitating a calibration dataset that reflects these variations. $\set{C}_\text{QC}$ is constructed using a {\em time step-aware sampling} method that captures the input distribution at arbitrary time steps, tailored to each baseline quantization method. 
First, PTQ4DM~\cite{ptq4dm} collects calibration data from a pre-trained full-precision (FP) model $\hat{\bfx}_\eta$ by using a Normally Distributed Time-step Calibration (NDTC) strategy, sampling $n_t$ images per time step $t$ with a Gaussian distribution centered on mid-time intervals to minimize degradation. 
In contrast, Q-Diffusion~\cite{q-diff} and TFMQ-DM~\cite{tfmq} employ uniform sampling across all time steps ($n_t = n_\text{fixed}$), maintaining a balance between dataset size and representational capacity. 
{\DPQ} adapts these time step-aware sampling strategies, constructing $\set{C}_\text{QC}$ to reflect the input distribution required for effective quantization while accommodating the characteristics of the base FP model.

\noindent \textbf{Distillation Calibration Dataset.} 
Distillation in {\DPQ} utilizes a calibration dataset $\set{C}_\text{DC}$ tailored to the student model, which has halved sampling steps compared to the quantized teacher model. 
To match this reduced step count, $\set{C}_\text{DC}$ is created using the full-precision (FP) model $\hat{\bfx}_\eta$, which provides a high-capacity reference for guiding the student model during distillation. This approach is especially critical when the teacher model is quantized, as the quantized teacher may lack sufficient capacity to effectively guide the student under traditional distillation methods. 
As shown in Fig.~\ref{fig:overall} (left), calibration data is generated at even time steps (e.g., $2,4,\ldots,T$) from the FP model with noise intensities closely matching those at the corresponding inference steps of the student model (e.g., $1,2,\ldots, T/2$), thus we call it {\em time-conditioned uniform sampling}. 
To introduce variability and avoid overfitting, random noise is applied at each step to generate stochastic, diverse outputs ($\set{C}_\text{DC} \leftarrow \set{C}_\text{DC} \cup \{\bfx_t^i\}$, where $i \sim \text{U}([1, n_t])$) at every step. 
By leveraging the FP model's stochastic outputs, $\set{C}_\text{DC}$ enables the student model to retain the generative quality and precision of the FP model, effectively mitigating limitations introduced by the quantized teacher. Details about $\set{C}_\text{DC}$ construction are provided in the Alg.~\ref{alg:Calibration Data}

\begin{algorithm}[h!]
\caption{Calibration Dataset Collection}
\label{alg:Calibration Data}
\begin{algorithmic}[1]
\State \textbf{Input:} Pre-trained diffusion model $\hat{\bfx}_\eta$, calibration datasets $\set{C_\text{QC}}, \set{C_\text{DC}}$, total sampling steps $T$, mean of the Normal distribution $\mu$, 
\State \textbf{Output:} Calibration datasets $\set{C_\text{QC}}, \set{C_\text{DC}}$

\vspace{-0.15cm}
\hspace{-1cm}
\hrulefill
\State Initialize $\set{C_\text{QC}} = \emptyset$, $\set{C_\text{DC}} = \emptyset$
\For{$t = 1, \ldots, T$}
    \State Sample $n_\text{max}$ intermediate inputs $\bfx_t^1, \ldots, \bfx_t^{\text{max}}$ from $\hat{\bfx}_\eta(\bfz_t^{n_\text{max}})$

    \State Determine $n_t$ as follows:
    \[
    n_t =
    \begin{cases} 
    n_t~\sim\mathcal{N}(\mu, T/2), & \text{if PTQ4DM}, \\
    n_\text{fixed}, & \text{else otherwise}
    \end{cases}
    \]
    
    \State Update $\set{C_\text{QC}} \gets \set{C_\text{QC}} \cup \{\bfx_t^1, \ldots, \bfx_t^{n_t}\}$

    
    \If{$t \mod 2 = 0$} \Comment{For distill calibration data}
    \State Update $\set{C_\text{DC}} \gets \set{C_\text{DC}} \cup \{\bfx_t^i\}$, where $i \sim \text{Uniform}([1, n_t])$
    \EndIf
\EndFor
\end{algorithmic}
\end{algorithm}

\subsection{Progressive Quantization (PQ)} \label{sec:PQ}

Quantizing DMs presents unique challenges, particularly at low precision (e.g., $4$-bit), where weight perturbations can lead to significant quantization noise~\cite{bit-shrinking}. 
Previous approaches~\cite{ptq4dm,q-diff,tfmq} primarily employed block or layer-wise reconstruction~\cite{brecq,adaptive_round} to minimize this noise, which proved effective for small calibration datasets and inter-layer dependencies but struggled with severe perturbations at lower precision levels.

\noindent \textbf{Two-stage Quantization.} 
To address this, our {\em Progressive Quantization (PQ)} framework follows a two-stage process to minimize errors during low-bit quantization. 
The first stage reduces the precision of a full-precision model $\hat{\bfx}_\eta$ to an intermediate bit-width $\tau$ 
to minimize early perturbations. 
In the second stage, the $\tau$-bit model $\hat{\bfx}_\eta^\tau$ undergoes further quantization to the target precision $\kappa$, producing the model $\hat{\bfx}_\eta^\kappa$. During init quantization, the scaling factors and clipping range of the weight quantizer $\set{Q}$ are set for an intermediate bit-width $\tau$, which is often chosen to be double the target bit-width $\kappa$ (i.e., $\tau = 2 \kappa$) for better memory alignment and efficient processing~\cite{towardsquant}.

\begin{algorithm}[t!]
\caption{Progressive Quantization (PQ)}
\label{alg:quant}
\begin{algorithmic}[1]
\State {\bfseries Input:} pre-trained FP model $\hat{\bfx}_\eta$ with weights $\eta$, quantization method $\set{Q}$, 
target bit-widths $\{\tau, \kappa\}$ with $\tau > \kappa$, calibration dataset $\set{C}_\text{QC}$, 
learning rate $\gamma$
\State {\bfseries Output:} quantized model $\hat{\bfx}_{\eta}^{\kappa}$

\vspace{-0.15cm}
\hspace{-1cm}
\hrulefill
\For{each block (or layer) $i = 1, 2, \dots, N$} 
    \State collect calibration outputs $\hat{\mathbf{a}}_i \in \set{C}_\text{QC}$ 

    \State set target bit-width: $b \leftarrow \tau$   
    \State initialize quantized weights $\eta_i^b = \set{Q}(\eta_i, b)$
    
    \State \texttt{/* weight quantization */}
    
    \For{each iteration $j = 1, 2, \dots $} 
        \State quantize block outputs to $b$-bits: $\mathbf{a}^b_i := \set{Q}(\hat{\mathbf{a}}_i, b)$
        
        \State compute perturbation: $\Delta \mathbf{a}_i = \mathbf{a}^b_i - \hat{\mathbf{a}}_i$

        \State update quantized weights using GD \eqref{eq:brecq-update}:
        $$\mathbf{\eta}_i^b \gets \mathbf{\eta}_i^b - \gamma \nabla_{\mathbf{\eta}_i^b} \mathbb{E} \Big[ \Delta \mathbf{a}_i^\top \mathbf{H}^{\mathbf{a}_i} \Delta \mathbf{a}_i \Big]$$

        \State round weights to $b$-bit: $\eta_i^b \gets \text{round}_b(\eta_i^b)$
        
        \If{bit-transition flag in Alg.~\ref{alg:momentum}} 
            \State break
        \EndIf
    \EndFor
\EndFor  

\State \texttt{/* activation quantization */}
\For{each block (or layer) $i = 1, 2, \dots, N$} 
    \State quantize activations to $b$-bits: $\mathbf{a}_i^b = \set{Q}(\hat{\mathbf{a}}_i, b)$
    \State update target bit-width: $b \leftarrow \kappa$ and repeat
\EndFor
\end{algorithmic}
\end{algorithm}

As described in Alg.~\ref{alg:quant}, for each block or layer $i$, the calibration outputs $\hat{\mathbf{a}}_i \in \set{C}_\text{QC}$ are quantized to a desirable bit-width $b$ (either $\tau$ or $\kappa$) using the quantizer $\set{Q}$: $\mathbf{a}_i^b := \set{Q}(\hat{\mathbf{a}}_i, b)$. Then, the perturbation $\hat{\mathbf{a}}_i$ is computed (line $9$), then used to minimize the objective \eqref{eq:block_recon}. The quantized model weights $\eta_i^b$ are iteratively refined by using the gradient update: 
\begin{equation} \label{eq:brecq-update}
    \mathbf{\eta}_i^b \gets \mathbf{\eta}_i^b - \gamma \nabla_{\mathbf{\eta}_i^b} \mathbb{E} \Big[ \Delta \mathbf{a}_i^\top \mathbf{H}^{\mathbf{a}_i} \Delta \mathbf{a}_i \Big],
\end{equation}
where $\gamma$ is the learning rate. This update step is repeated to ensure the quantized weights align closely with the FP model, reducing reconstruction errors. 
Unlike weight quantization, activation quantization is more challenging due to the variability across time steps during inference. To address this, activation quantization is performed after weight quantization (lines $19$-$22$) to facilitate smoother optimization and effectively minimize reconstruction error. 
Once sufficient optimization is achieved at $\tau$ bit-width, the transition to the second stage (for target bit-width $\kappa$) is dynamically determined using a bit-transition detection mechanism. 

\smallskip
\noindent \textbf{Momentum-based Bit Transition Detection.} \label{sec:momentum}
In PQ, the timing of bit-width transitions is crucial. A rapid transition to the target bit-width $\kappa$ fails to leverage the benefits of higher bit-width $\tau$, leading to inadequate weight perturbation reduction and generative quality degradation. 
Conversely, overly delayed transitions hamper optimization at the target precision $\kappa$, causing reconstruction errors. 
Prior work~\cite{bit-shrinking} introduced adaptive bit-width reductions with fixed cycles to minimize perturbation. However, the large scale and variability of DMs make frequent transitions ineffective, as they leave insufficient time for proper optimization of perturbation errors. 


\begin{algorithm}[t!]
\caption{Momentum-based Bit Transition Detection}
\label{alg:momentum}
\begin{algorithmic}[1]
\State {\bfseries Input:} perturbation loss window $\set{W}_\text{pert}$ of size $W$(queue), 
momentum coefficient $\beta$, threshold $\pi$, 
small constant $\epsilon$

\State {\bfseries Output:} bit-transition flag (True/False)
\State {\bfseries Initialize:} perturbation loss window $\set{W}_\text{pert} = \emptyset$, $\mathcal{G} = 0$

\vspace{-0.15cm}
\hspace{-1cm}
\hrulefill

    \State compute perturbation: $\Delta \mathbf{a}_i = \mathbf{a}_i^\tau - \hat{\mathbf{a}}_i$

    \State Append to loss window: $\set{W}_\text{pert} \gets \set{W}_\text{pert} \cup \Delta \mathbf{a}_i$
    
    
    \If{$\mathcal{G} = 0$}
        set $\mathcal{G} \gets \Delta \mathbf{a}_i$
    \Else
        ~update momentum: $\mathcal{G} \gets \beta \cdot \mathcal{G} + (1 - \beta) \cdot \Delta \mathbf{a}_i$
    \EndIf

    \If{$|\set{W}_\text{pert}| = W$}
        \State compute average loss and momentum change rate: 
        $$L_\text{avg} \gets \frac{1}{W} \sum_{l \in \set{W}_\text{pert}} l, ~~ \text{rate} \gets \frac{|L_\text{avg} - \set{G}|}{\max(L_\text{avg}, \epsilon)} $$        
        \If{$\text{rate} < \pi$}
             return \textbf{True} (i.e., transition detected)
        \EndIf
       
    \EndIf
    
\end{algorithmic}
\end{algorithm}
To address this, we propose a {\em momentum-based bit transition detection} to detect the optimal point of transition while using a fixed intermediate bit-width $\tau$. 
This mechanism identifies steep gradients in weight perturbation loss at $\tau$-bit model, signaling sufficient optimization and readiness to transition to $\kappa$. The momentum $\set{G}$, accumulates gradients to track changes in perturbation, is updated as follows: 
\begin{equation} \label{eq:momentum}
    \mathcal{G} \gets \beta \cdot \mathcal{G} + (1 - \beta) \cdot \Delta \mathbf{a}_i,
\end{equation}
where $\beta \in [0,1)$ is the momentum coefficient, and $\Delta \mathbf{a}_i$ represents the change in perturbation for the $i$-th block or layer. 
When the average change rate of perturbation loss (computed within a fixed-size window) falls below a threshold $\pi$, the bit-width transitions to $\kappa$, as details are provided Alg.~\ref{alg:momentum}. 
This ensures transitions occur only when higher-bit optimization saturates, preventing premature or delayed transitions that could degrade performance. 

\vspace{-0.1cm}
\subsection{Calibration-Assisted Distillation (CAD)
}
While diffusion distillation~\cite{progressivedistill} has been proposed to accelerate inference for full-precision (FP) DMs, it struggles to preserve the student's precision in quantized environments due to its reliance solely on a teacher-dependent inheritance structure. 
To address this limitation, we propose Calibration-Assisted Distillation (CAD), which minimizes errors during the distillation process of a quantized teacher model $\hat{\bfx}_\eta^\kappa$ by incorporating a dedicated calibration dataset $\set{C}_\text{DC}$. 
CAD builds upon conventional distillation~\cite{knowledgedistill}, which typically combines two loss terms: the first term for classification performance between the ground-truth and the student network, and the second term for the difference between the teacher and student networks. 
In contrast, diffusion distillation~\cite{progressivedistill} employs only the second term, leading to weaker precision for the quantized student network $\hat{\bfx}_\theta^\kappa$. 
To overcome this, we introduce a calibration dataset $\set{C}_\text{DC}$ generated by the FP model as the ground-truth, enabling a direct comparison with the student. 
\begin{algorithm}[t]
\caption{Calibration-Assisted Distillation (CAD)}
\label{alg:CAD}
\begin{algorithmic}[1]
\State {\bfseries Input:} quantized teacher DM $\hat{\bfx}^\kappa_\eta$ with weights $\eta$ and sampling steps $T$, training dataset $\set{D}$, calibration dataset $\set{C}_\text{DC}$, noise schedule factors $\{\alpha_t, \sigma_t\}_{t=0}^T$, coefficient function $w(\cdot)$, hyperparameter $\lambda$, learning rate $\gamma$

\State {\bfseries Output:} distilled student DM $\hat{\bfx}_{\theta}^{\kappa}$ with weights $\theta$

\vspace{-0.15cm}
\hspace{-1cm}
\hrulefill

\State {\bfseries initialize:} $\hat{\bfx}_{\theta}^{\kappa} = \hat{\bfx}_{\eta}^{\kappa}$ 
\Comment{copy student from teacher}

\While{not converged} \Comment{distillation}
    \State sample data and noise: $\bfx \sim \set{D}$, $\bmeps \sim \set{N}({\bm 0}, \textbf{I})$

    \State sample time step $t = i/T$ with $i \sim \text{Cat}[1,2,\ldots,T]$ 

    \State add noise to data: $\bfz_t = \alpha_t \bfx + \sigma_t \bmeps$

    \State compute intermediate steps: 
    $$t' = t - 0.5/T, \quad t'' = t - 1/T$$

    \State perform $2$ steps of DDIM sampling with $\hat{\bfx}_{\eta}^{\kappa}$:
    \begin{align*}
        \bfz_{t'} &= \alpha_{t'} \hat{\bfx}_{\eta}^{\kappa}(\bfz_t) + \frac{\sigma_{t'}}{\sigma_t} \big( \bfz_t - \alpha_t \hat{\bfx}_{\eta}^{\kappa}(\bfz_t) \big), \cr 
    \bfz_{t''} &= \alpha_{t''} \hat{\bfx}_{\eta}^{\kappa}(\bfz_{t'}) + \frac{\sigma_{t''}}{\sigma_{t'}} \big( \bfz_{t'} - \alpha_{t'} \hat{\bfx}_{\eta}^{\kappa}(\bfz_{t'}) \big) 
    \end{align*}

    \State compute the teacher target: $\tilde{\bfx} = \frac{\bfz_{t''} - (\sigma_{t''}/\sigma_t) \bfz_t }{ \alpha_{t''} - (\sigma_{t''}/\sigma_t) \alpha_t }$

    \State sample calibration data: $\bfx_t \sim \{ \bfx_t^i \mid \bfx_t^i \in \set{C}_\text{DC} \}$
    
    \State update student model via SGD with $\set{L}_\text{CAD}$ in \eqref{eq:CAD} as: 
    \begin{align*}
        \set{L}_\text{CAD} &= w(\lambda_t) \| \tilde{\bfx} - \hat{\bfx}_{\theta}^{\kappa}(\bfz_t) \|_2^2+\lambda \cdot \text{EM}(\bfx_t,\hat{\bfx}^\kappa_{\theta}(\bfz_t)), \cr 
        \theta &\gets \theta - \gamma \nabla_{\theta} \set{L}_\text{CAD}
    \end{align*}
\EndWhile

\end{algorithmic}
\end{algorithm}

\begin{table*}[h!]
\centering
\resizebox{\textwidth}{!}{%
\begin{tabular}{lcccccccccc}
\toprule
\multirow{2}{*}{\textbf{Methods}} & \multirow{2}{*}{\textbf{Time Step}} & \multirow{2}{*}{\textbf{Bits (W/A)}} & \multicolumn{2}{c}{\textbf{LSUN-Bedrooms}} & \multicolumn{2}{c}{\textbf{LSUN-Churches}} & \multicolumn{2}{c}{\textbf{CelebA-HQ }} & \multicolumn{2}{c}{\textbf{CIFAR-10}} \\
\cmidrule(lr){4-5} \cmidrule(lr){6-7} \cmidrule(lr){8-9} \cmidrule(lr){10-11}
 &  &  & FID$\downarrow$ & sFID$\downarrow$ & FID$\downarrow$ & sFID$\downarrow$ & FID$\downarrow$ & sFID$\downarrow$ & FID$\downarrow$ & IS$\uparrow$ \\
\midrule
Full Prec.  & 100 & 32/32 & 2.98 & 7.09 & 4.12 & 10.89 & 8.74 & 10.16 & 4.23 & 9.04 \\
Distill DM only & 50 & 32/32 & 3.07 & 7.15 & 4.26 & 11.01 & 9.01& 10.83 & 4.28 & 9.12 \\ 
\midrule
PTQ4DM & 100 & 8/8 & 4.75 & 9.59 & 4.80 & 13.48 & 14.42 & 15.06 & 19.59 & 9.02 \\
Q-Diff & 100 & 8/8 & 4.51 & 8.17 & 4.41 & 12.23 & 12.85 & 14.16 & 4.78 & 8.82 \\
TFMQ-DM & 100 & 8/8 & 3.14 & 7.26 & 4.01 & 10.98 & 8.91 & 10.20 & 4.24 & 9.07 
\\ 
\cdashline{1-11}[2pt/1pt] \vspace{-0.3cm}\\
PTQ4DM \textbf{+ Bit-shrink} & 100 & 8/8 & 6.45 & 10.75 & 7.68 & 15.49 & 18.12 & 19.69 & 22.37 & 8.08 \\
Q-Diff \textbf{+ Bit-shrink}& 100 & 8/8 & 6.48 & 10.15 & 7.44 & 15.31 & 18.01 & 19.75 & 8.29 & 8.11 \\
TFMQ-DM \textbf{+ Bit-shrink}& 100 & 8/8 & 5.95& 10.11 & 7.25 & 14.97 & 18.43 & 18.46 & 8.62 & 7.85 
\\ 
\hdashline \vspace{-0.3cm}\\
PTQ4DM \textbf{+ PQ (Ours)} & 100 & 8/8 & 4.26 & 9.12 & 4.22 & 12.84 & 14.01 & 14.70 & 13.83 & 9.10 \\
Q-Diff \textbf{+ PQ (Ours)}& 100 & 8/8 & 4.15 & 7.63 & 4.12 & 11.98 & 12.25 & 13.74 & 4.39 & 8.99 \\
TFMQ-DM \textbf{+ PQ (Ours)}& 100 & 8/8 & 3.06& 7.07 & 3.86 & 10.89 & 8.57 & 10.11 & 4.12 & 9.61 
\\ 
\midrule
PTQ4DM \textbf{+ Distill-DM.} & 50 & 8/8 & 7.65 & 11.95 & 6.05 & 15.96 & 16.19 & 16.47 & 21.46& 7.41 \\
Q-Diff \textbf{+ Distill-DM.}& 50 & 8/8 & 6.12 & 10.75 & 5.98 & 14.85 & 14.10 & 15.76 & 5.69 & 7.12 \\
TFMQ-DM \textbf{+ Distill-DM.}& 50 & 8/8 & 4.97 & 9.55 & 5.92 & 14.08 & 13.75 & 15.49 & 5.59 & 7.36
\\ 
\hdashline \vspace{-0.3cm}\\
PTQ4DM \textbf{+ CAD (Ours)} & 50 & 8/8 & 5.82 & 10.24 & 5.21 & 13.94 & 14.68 & 15.40 & 17.65 & 9.11 \\
Q-Diff \textbf{+ CAD (Ours)}& 50 & 8/8 & 4.67 & 8.25 & 4.54 & 12.33 & 12.73 & 14.22 & 4.60 & 9.00 \\
TFMQ-DM \textbf{+ CAD (Ours)}& 50 & 8/8 & 3.27 & 7.36 & 4.18 & 11.12 & 9.01 & 11.46 & 4.33 & 9.21 \\
\cdashline{1-11}[2pt/1pt] \vspace{-0.3cm}\\
\textbf{\DPQ~(PTQ4DM)} &\cellcolor[HTML]{e6e6e6}50 &\cellcolor[HTML]{e6e6e6} 8/8 &\cellcolor[HTML]{e6e6e6} \textbf{4.56} (\textcolor{blue}{\textit{-0.19}}) &\cellcolor[HTML]{e6e6e6} \textbf{9.40} (\textcolor{blue}{\textit{-0.19}}) &\cellcolor[HTML]{e6e6e6} \textbf{4.61} (\textcolor{blue}{\textit{-0.19}}) & \cellcolor[HTML]{e6e6e6}\textbf{13.12} (\textcolor{blue}{\textit{-0.36}}) &\cellcolor[HTML]{e6e6e6} \textbf{14.24} (\textcolor{blue}{\textit{-0.18}}) &\cellcolor[HTML]{e6e6e6} \textbf{15.03} (\textcolor{blue}{\textit{-0.03}}) &\cellcolor[HTML]{e6e6e6} \textbf{14.12} (\textcolor{blue}{\textit{-5.47}}) &\cellcolor[HTML]{e6e6e6} \textbf{9.05} (\textcolor{blue}{\textit{+0.03}}) \\ 
\textbf{\DPQ~(Q-Diff)} & \cellcolor[HTML]{e6e6e6}50 &\cellcolor[HTML]{e6e6e6} 8/8 &\cellcolor[HTML]{e6e6e6} \textbf{4.36} (\textcolor{blue}{\textit{-0.15}}) &\cellcolor[HTML]{e6e6e6} \textbf{7.88} (\textcolor{blue}{\textit{-0.29}}) &\cellcolor[HTML]{e6e6e6} \textbf{4.29} (\textcolor{blue}{\textit{-0.12}}) & \cellcolor[HTML]{e6e6e6}\textbf{12.10} (\textcolor{blue}{\textit{-0.13}}) &\cellcolor[HTML]{e6e6e6} \textbf{12.56} (\textcolor{blue}{\textit{-0.29}}) & \cellcolor[HTML]{e6e6e6}\textbf{13.93} (\textcolor{blue}{\textit{-0.23}}) &\cellcolor[HTML]{e6e6e6} \textbf{4.53} (\textcolor{blue}{\textit{-0.25}}) & \cellcolor[HTML]{e6e6e6}\textbf{8.91} (\textcolor{blue}{\textit{+0.02}})\\
\textbf{\DPQ~(TFMQ-DM)} &\cellcolor[HTML]{e6e6e6} 50 &\cellcolor[HTML]{e6e6e6} 8/8 &\cellcolor[HTML]{e6e6e6} \textbf{3.10} (\textcolor{blue}{\textit{-0.04}}) &\cellcolor[HTML]{e6e6e6} \textbf{7.12} (\textcolor{blue}{\textit{-0.14}}) &\cellcolor[HTML]{e6e6e6} \textbf{3.92} (\textcolor{blue}{\textit{-0.09}}) &\cellcolor[HTML]{e6e6e6} \textbf{10.94} (\textcolor{blue}{\textit{-0.04}}) &\cellcolor[HTML]{e6e6e6} \textbf{8.65} (\textcolor{blue}{\textit{-0.26}}) &\cellcolor[HTML]{e6e6e6} \textbf{10.15} (\textcolor{blue}{\textit{-0.05}}) &\cellcolor[HTML]{e6e6e6} \textbf{4.19} (\textcolor{blue}{\textit{-0.05}}) &\cellcolor[HTML]{e6e6e6} \textbf{9.53} (\textcolor{blue}{\textit{+0.46}}) \\ 
\midrule
PTQ4DM & 100 & 4/8 & 20.72 & 54.30 & 4.97 & 14.87 & 17.08 & 17.48 & 5.14 & 8.93 \\
Q-Diff & 100 & 4/8 & 6.40 & 17.93 & 4.66 & 13.94 & 15.55 & 16.86 & 4.98 & 9.12 \\
TFMQ-DM & 100 & 4/8 & 3.68 & 7.65 & 4.14 & 11.46 & 8.76 & 10.26 & 4.78 & 9.13 \\ 
\cdashline{1-11}[2pt/1pt] \vspace{-0.3cm}\\
\textbf{\DPQ~(PTQ4DM)} &\cellcolor[HTML]{e6e6e6} 50 &\cellcolor[HTML]{e6e6e6} 4/8 & \cellcolor[HTML]{e6e6e6}\textbf{20.47} (\textcolor{blue}{\textit{-0.25}}) &\cellcolor[HTML]{e6e6e6} \textbf{39.88} (\textit{-14.42}) &\cellcolor[HTML]{e6e6e6} \textbf{4.91} (\textcolor{blue}{\textit{-0.06}}) &\cellcolor[HTML]{e6e6e6} \textbf{14.02} (\textcolor{blue}{\textit{-0.85}}) &\cellcolor[HTML]{e6e6e6} \textbf{16.89} (\textcolor{blue}{\textit{-0.19}}) &\cellcolor[HTML]{e6e6e6} \textbf{17.09} (\textcolor{blue}{\textit{-0.39}}) &\cellcolor[HTML]{e6e6e6} \textbf{4.98} (\textcolor{blue}{\textit{-0.16}}) &\cellcolor[HTML]{e6e6e6} \textbf{9.05} (\textcolor{blue}{\textit{+0.12}}) \\ 
\textbf{\DPQ~(Q-Diff)} &\cellcolor[HTML]{e6e6e6} 50 & \cellcolor[HTML]{e6e6e6} 4/8 & \cellcolor[HTML]{e6e6e6}\textbf{6.29} (\textcolor{blue}{\textit{-0.11}}) &\cellcolor[HTML]{e6e6e6} \textbf{16.12} (\textcolor{blue}{\textit{-1.81}}) &\cellcolor[HTML]{e6e6e6} \textbf{4.59} (\textcolor{blue}{\textit{-0.07}}) &\cellcolor[HTML]{e6e6e6} \textbf{12.65} (\textcolor{blue}{\textit{-1.29}}) &\cellcolor[HTML]{e6e6e6} \textbf{15.01} (\textcolor{blue}{\textit{-0.54}}) &\cellcolor[HTML]{e6e6e6} \textbf{16.20} (\textcolor{blue}{\textit{-0.66}}) & \cellcolor[HTML]{e6e6e6}\textbf{4.82} (\textcolor{blue}{\textit{-0.16}}) &\cellcolor[HTML]{e6e6e6} \textbf{9.17} (\textcolor{blue}{\textit{+0.05}}) \\
\textbf{\DPQ~(TFMQ-DM)} & \cellcolor[HTML]{e6e6e6} 50 & \cellcolor[HTML]{e6e6e6} 4/8 &\cellcolor[HTML]{e6e6e6} \textbf{3.64} (\textcolor{blue}{\textit{-0.04}}) &\cellcolor[HTML]{e6e6e6} \textbf{7.46} (\textcolor{blue}{\textit{-0.19}}) &\cellcolor[HTML]{e6e6e6} \textbf{3.82} (\textcolor{blue}{\textit{-0.32}}) &\cellcolor[HTML]{e6e6e6} \textbf{11.44} (\textcolor{blue}{\textit{-0.02}}) &\cellcolor[HTML]{e6e6e6} \textbf{8.57} (\textcolor{blue}{\textit{-0.19}}) &\cellcolor[HTML]{e6e6e6} \textbf{10.12} (\textcolor{blue}{\textit{-0.14}}) &\cellcolor[HTML]{e6e6e6} \textbf{4.73} (\textcolor{blue}{\textit{-0.05}}) &\cellcolor[HTML]{e6e6e6} \textbf{9.15} (\textcolor{blue}{\textit{+0.02}}) \\
\bottomrule
\end{tabular}%
}

\begin{tikzpicture}[overlay, remember picture]
    \draw[thick,<->,black] (-8.9, 4.5) .. controls (-9.1, 4.8) and (-9.1, 5.2) .. (-8.9, 5.5) node[midway, left] {\footnotesize \textcircled{\scriptsize 2}};
     \draw[thick,<->,black] (-8.9, 2.8) .. controls (-9.1, 3.1) and (-9.1, 3.5) .. (-8.9, 3.8) node[midway, left] {\footnotesize \textcircled{\scriptsize 3}};
     \draw[thick,<->,black] (9, 2.2) .. controls (9.2, 2.8) and (9.2, 5.7) .. (9, 6.3) node[midway, right] {\footnotesize \textcircled{\scriptsize 1}};
     \draw[thick,<->,black] (9, 1.2) .. controls (9.2, 0.9) and (9.2, 0.5) .. (9, 0.2) node[midway, right] {\footnotesize \textcircled{\scriptsize 1}};
\end{tikzpicture}
\caption{
Performance comparisons of {\DPQ} with baseline methods across various datasets and metrics (FID, sFID, IS). Results are shown for different quantization settings (Bits (W/A)) and sampling steps ($T$). Improvements over baselines are highlighted in \textcolor{blue}{blue}.
}
\label{tb:DPQ_all_results}
\end{table*}

As illustrated in Fig.~\ref{fig:overall} (right), CAD introduces a dual-objective loss function that combines knowledge distillation loss ($Loss_\text{KD}$) in \cite{progressivedistill} and calibration loss ($Loss_\text{CD}$). This ensures that the student model $\hat{\bfx}_\theta^\kappa$ inherits knowledge from the teacher $\hat{\bfx}_\eta^\kappa$ while aligning its outputs with the calibration dataset obtained from the FP model. The objective function is defined as:
\begin{equation} \label{eq:CAD}
    \set{L}_\text{CAD} = 
    w(\lambda_t) \underbrace{ \| \tilde{\bfx} - \hat{\bfx}_{\theta}^{\kappa}(\bfz_t) \|_2^2 }_{Loss_\text{KD}} 
    + \lambda \cdot \underbrace{\text{EM}(\set{C_\text{DC}}, \hat{\bfx}_{\theta}^{\kappa}(\bfz_t))}_{Loss_\text{CD}},
\end{equation}
where $\tilde{\bfx}$ represents the teacher model's output after two sampling steps and $w(\lambda_t)$ is a pre-defined coefficient function proposed by diffusion distillation that adjusts loss function weights at each time step $t$ (see \eqref{eq:loss_DM} and \eqref{eq:distill-DM}).
$\text{EM}(\cdot)$ denotes the Earth-Mover Distance, which measures the minimal effort required to align two distributions, making it highly sensitive to small differences, and $\lambda$ is a hyperparameter to balance the contributions of $Loss_\text{KD}$ and $Loss_\text{CD}$. 
Note that the calibration loss $Loss_\text{CD}$ computes the EM distance between the student model's output $\hat{\bfx}_\theta^\kappa(\bfz_t)$ at the current time step (e.g., $T/2$) and the FP calibration output from the corresponding time step (e.g., $T$). By combining two loss terms, CAD ensures both knowledge transfer from the teacher and calibration consistency with the FP model. Please refer to Alg.~\ref{alg:CAD} for implementation details.

\vspace{-0.2cm}
\section{Experiments}

\subsection{Experiments Setup.} 
To evaluate the effectiveness of {\DPQ}, we conducted extensive experiments across multiple generative benchmarks, emphasizing improvements in generative performance, computational efficiency, and compression quality.  

\noindent {\bf Datasets and Baselines.} 
We evaluated {\DPQ} on CIFAR-10 ($32 \times 32$), ImageNet ($64 \times 64$), CelebA-HQ and LSUN-Bedrooms/Churches ($256 \times 256$ high-resolution) datasets. 
Two prominent diffusion model (DM) architectures, DDIM~\cite{ddim} and Latent Diffusion Models (LDM)~\cite{ldm}, were employed. 
{\DPQ} was implemented with SOTA quantization baselines $\set{Q}$: (i) PTQ4DM~\cite{ptq4dm}, employing block-wise reconstruction and QDrop; (ii) Q-Diff~\cite{q-diff}, using layer-wise reconstruction and split quantization; and TFMQ-DM~\cite{tfmq}, an extension of Q-Diff with temporal-aware calibration. 
Quantization calibration datasets $\set{C}_\text{QC}$ were tailored to specific methods: constructed using NDTC for PTQ4DM, and time-step specific and batch-sized sampling for Q-Diff and TFMQ-DM. Distillation calibration dataset $\set{C}_\text{DC}$ was created from FP models, with a size equal to the number of time steps $T$.

\noindent {\bf Metrics and Experimental Configurations.} 
Generative quality is assessed using Frechet Inception Distance (FID), sliding FID (sFID), and Inception Score (IS) on $50$K generated images. 
These images are also evaluated for downstream classification tasks (Precision, Recall, and F1 score). 
Experiments were conducted with various bit-widths, where Bit(W/A) denotes the precision of weights/activations. Experiments use $T=100$ sampling steps for baselines and $50$ steps for {\DPQ}. 

For progressive quantization, we set bit-width sets $\tau$ and $\kappa$ to differ by a factor of two. Quantized weights are reconstructed over 20K iterations with a batch size of 32 for both DDIM and LDM. For activation quantization, PTQ4DM adopts QDrop~\cite{qdrop}, Q-Diff uses step size learning~\cite{learnedstepsize}, and TFMQ-DM estimates activation ranges via EMA~\cite{tfmq}. For momentum-based bit transition, we use a threshold $\pi = 0.04$, momentum coefficient $\beta = 0.9$, and $\epsilon = 1\mathrm{e}{-8}$ to prevent numerical instability. The loss window size $W$ is set to 500.

\begin{figure*}[t!]
  \centering{\includegraphics[width=\textwidth]{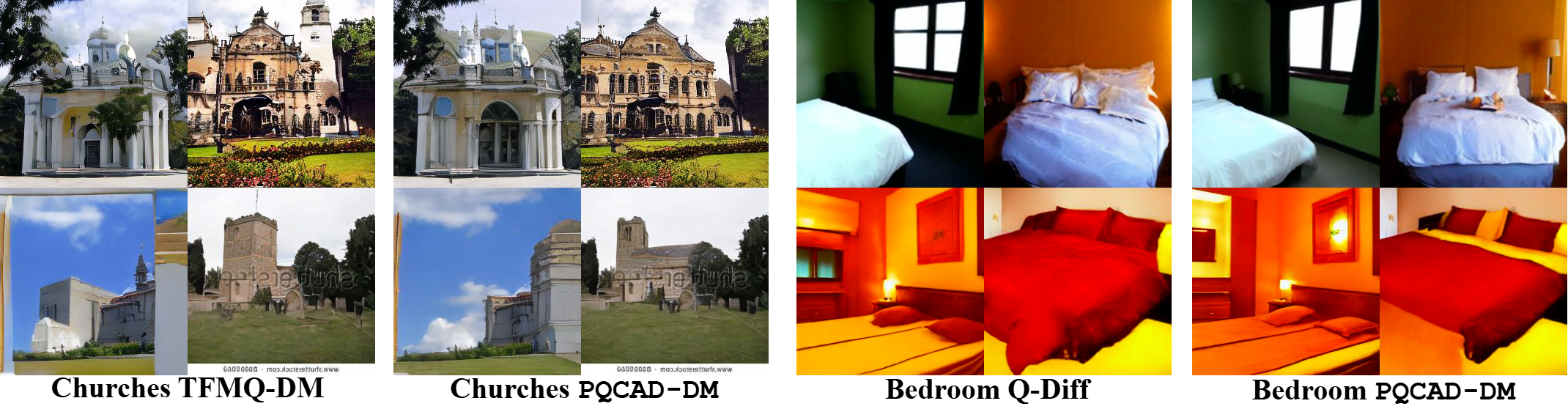}}
  \vspace{-0.2cm}
  \caption{ $256 \times 256$ image generation results using TFMQ-DM, Q-Diff and {\DPQ} under $4/8$-bit precision.
}
  \label{fig:image_churches}
  \vspace{-0.25cm}
\end{figure*}

\subsection{Main Results}

\noindent \textbf{Comparison with Baselines.} As shown in Tab.~\ref{tb:DPQ_all_results}~\textcircled{\scriptsize 1}, {\DPQ} demonstrates superior generative quality and computational efficiency compared to three popular quantization methods: PTQ4DM, Q-Diff, and TFMQ-DM. 
{\DPQ} halves the sampling steps while maintaining or surpassing baseline performance across all datasets. 
For example, on CelebA-HQ dataset, {\DPQ} with TFMQ-DM achieves an FID of $8.57$ and sFID of $10.12$ at $50$ sampling steps, outperforming TFMQ-DM's FID of $8.76$ and sFID of $10.26$ at $100$ steps, and even matching the full-precision model (FID of $8.74$ and sFID of $10.16$). 
Similar improvements are observed on LSUN-Bedrooms/Churches and CIFAR-10, where {\DPQ} consistently improves FID, sFID and IS metrics under various quantization scenarios. 
Notably, {\DPQ} excels under strict compression settings, such as $4/8$-bit quantization. On LSUN-Churches, it achieves an FID of $3.64$ and sFID of $7.46$, showcasing significant computational efficiency. Fig.~\ref{fig:image_churches} illustrates {\DPQ}'s qualitative superiority, showcasing better preservation of fine-grained details and structural consistency compared to TFMQ-DM. 

\begin{figure}[t!]
\centering
\includegraphics[width=0.8\linewidth]{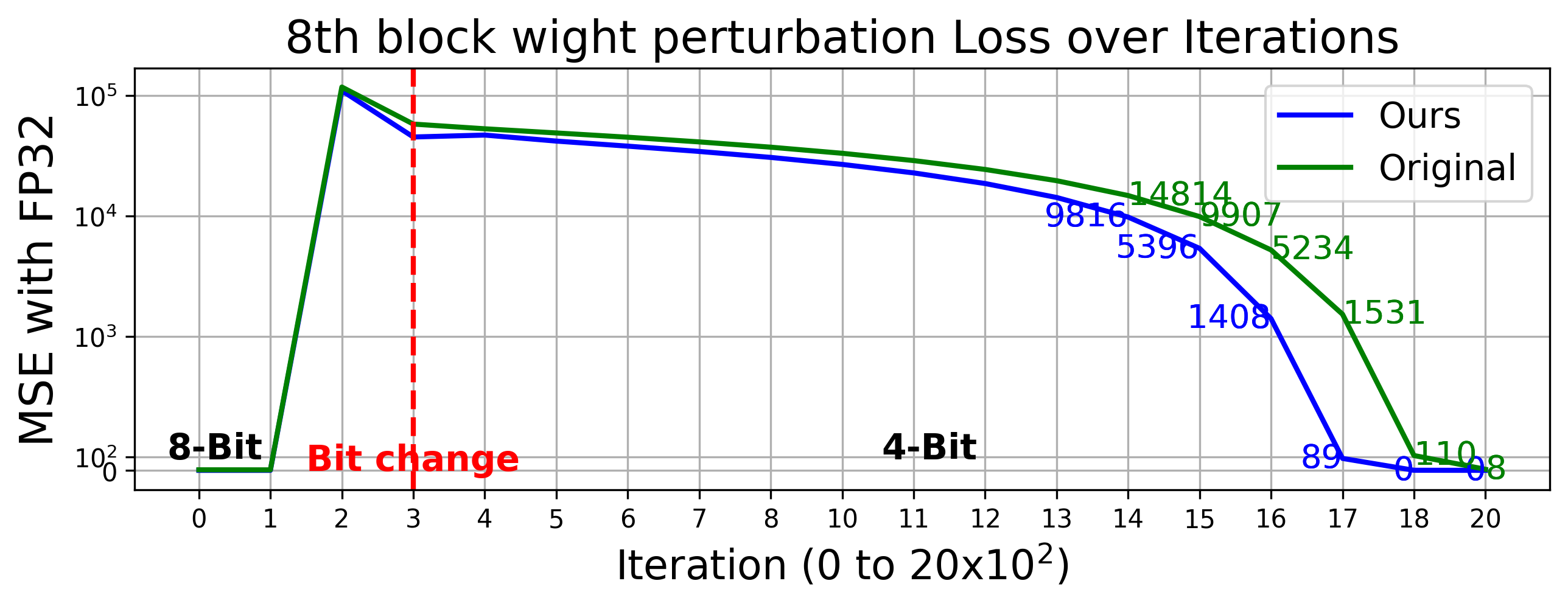}
\resizebox{0.9\linewidth}{!}{%
\begin{tabular}{lccc}
\toprule
\multirow{3}{*}{\textbf{Methods}} & \multicolumn{3}{c}{G-Mean Perturb ($\downarrow$) \& $\Delta_{\text{Pert}}\text{(orig$-$ours)} (\uparrow)$} \\
\cmidrule(lr){2-4} 
& \textbf{Bedroom} & \textbf{Churches} & \textbf{CelebA} \\
\midrule
Q-Diffusion   & 21.75 & 22.14 & 37.40 \\
\textbf{+ PQ (ours)} &\bf{ 16.26 ($\Delta 311$)}& \bf{17.62 ($\Delta 341$)} & \bf{32.95 ($\Delta 367$)} \\
\cdashline{1-4}[2pt/1pt]\vspace{-0.3cm} \\
PTQ4DM & 25.79 & 24.93 & 41.55 \\
\textbf{+ PQ (ours)} & \bf{21.58 ($\Delta 338$)} & \bf{20.76 ($\Delta 377$)} &\bf{ 34.06 ($\Delta 383$)} \\
\cdashline{1-4}[2pt/1pt] \vspace{-0.3cm} \\
TFMQ-DM & 19.94 & 19.42 & 32.41 \\
\textbf{+ PQ (ours)} & \bf{16.70 ($\Delta 294$)} & \bf{15.57 ($\Delta 352$)} & \bf{28.97 ($\Delta 309$)} \\
\bottomrule
\end{tabular}%
}
\caption{(Top) Weight perturbation loss over iterations for the $8$th block, comparing TFMQ-DM at $4$-bit with {\DPQ} transitioning from $8$ to $4$-bit. 
(Bottom) Geometric mean perturbation (G-Mean) and improvement ($\Delta_\text{Pert}$) across methods quantized to $4/8$ bits. 
}
\label{fig:combined}

\end{figure}

\noindent \textbf{Impact of PQ.} The effectiveness of PQ is highlighted in Tab.~\ref{tb:DPQ_all_results}~\textcircled{\scriptsize 2}. 
Momentum-based bit transitions in PQ consistently outperform frequent bit-shrinking methods~\cite{bit-shrinking}. 
For instance, on CIFAR-10, PQ with $8/8$-bit PTQ4DM reduces the FID from $19.59$ to $13.83$ and improves IS from $9.02$ to $9.10$ compared to PTQ4DM alone. 
PQ effectively minimizes perturbations at higher bit-widths ($\tau$) and transitions smoothly to lower bit-widths ($\kappa$), as indicated by the geometric mean perturbation improvement ($\Delta_{\text{Pert}}$) across datasets (see Fig.~\ref{fig:combined} (Bottom)). 
Additionally, {\DPQ}'s adaptive bit transition mechanism significantly reduces perturbations after transitioning, as seen in Fig.~\ref{fig:combined} (Top), where the optimization process accelerates across all blocks due to the stimulating effect of momentum-based transitions.

\noindent \textbf{Impact of CAD.} Tab.~\ref{tb:DPQ_all_results}~\textcircled{\scriptsize 3} highlights CAD's ability to mitigate distillation-induced errors. 
CAD significantly improves FID and sFID metrics compared to traditional Distill-DM~\cite{progressivedistill}. 
For example, on LSUN-Churches, CAD applied to Q-Diff reduces an FID from $5.98$ to $4.54$ and sFID from $14.85$ to $12.33$, compared to Q-Diff with Distill-DM. 
Even with halved sampling steps, {\DPQ} achieves performance comparable to non-distilled Q-Diff (FID $4.41$ and sFID $12.23$). Similar trends are observed across other base methods, underscoring CAD's robustness in maintaining generative quality under reduced precision. 

\noindent \textbf{Computational Efficiency Analysis.} We evaluate the model size, inference time, throughput, and maximum batch size and bit-operations (GBOPs) across multiple datasets in Table~\ref{tab:throughput_all}. All models operate under identical GPU A6000 memory constraints. GBOPs are computed as the total bit-level operations per forward pass, considering both weights and activations.
{\DPQ} consistently achieves faster inference, higher throughput, and larger batch sizes than both Q-Diff and TFMQ-DM. For instance, on LSUN-Bedroom, {\DPQ} runs 1.6× faster than Q-Diff (204.65 vs. 340.11 minutes) while achieving double the efficiency in terms of GBOPs (3,240 vs. 6,480). Similar trends are observed across other datasets. By aggressively reducing bit-widths without degrading generation quality, {\DPQ} significantly lowers computational overhead, enabling superior hardware utilization in resource-constrained environments.

\begin{table*}[t!]
\centering
\large
\resizebox{0.85\textwidth}{!}{
\begin{tabular}{lcccccc}
\toprule
\textbf{Dataset} & \textbf{Method} & \textbf{Model Size (MB)} & \textbf{Time (min)} & \textbf{Throughput (img/sec)} & \textbf{Max Batch Size} & \textbf{GBOPs} \\
\midrule
\multirow{4}{*}{LSUN-Churches} 
    & FP32    & 1,179.9 & 5,655.7 & 0.013 & 1.0× & 22,170 \\
    & Q-Diff    & 147.5 & 342.15 & 0.224 & 2.6× & 1,340 \\
    & TFMQ-DM   & 147.5 & 315.43 & 0.264 & 3.2× & 1,340 \\
    & \textbf{\DPQ} & \textbf{147.5}  & \textbf{206.52} & \textbf{0.440} & \textbf{3.2×} & \textbf{670} \\
\cmidrule{1-7}
\multirow{4}{*}{LSUN-Bedroom} 
    & FP32    & 1,096.2 & 5,620.33 & 0.013 & 1.0× & 107,170 \\
    & Q-Diff    & 137 & 340.11 & 0.226 & 2.4× & 6,480 \\
    & TFMQ-DM   & 137 & 312.39 & 0.267 & 2.9× & 6,480 \\
    & \textbf{\DPQ} & \textbf{137}  & \textbf{204.65} & \textbf{0.446} & \textbf{2.9×} & \textbf{3,240} \\
\cmidrule{1-7}
\multirow{4}{*}{CelebA} 
    & FP32    & 1,096.2 & 5,571.52 & 0.014 & 1.0× & 107,170 \\
    & Q-Diff    & 137 & 336.97 & 0.232 & 2.4× & 6,480 \\
    & TFMQ-DM   & 137 & 308.45 & 0.271 & 2.9× & 6,480 \\
    & \textbf{\DPQ} & \textbf{137}  & \textbf{202.16} & \textbf{0.454} & \textbf{2.9×} & \textbf{3,240} \\
\cmidrule{1-7}
\multirow{4}{*}{CIFAR-10} 
    & FP32   & 143.2  & 505.82  & 0.164 & 1.0× & 6,597 \\
    & Q-Diff    & 17.9  & 30.60  & 2.723 & 2.6× & 399 \\
    & TFMQ-DM   & 17.9  & 27.01  & 3.086 & 3.0× & 399 \\
    & \textbf{\DPQ} & \textbf{17.9}  & \textbf{17.44} & \textbf{4.790} & \textbf{3.0×} &  \textbf{199.5} \\
\bottomrule
\end{tabular}
}
\caption{
Comparison of model size, inference time, throughput (images/sec), max batch size, and GBOPs (bit-operations) for 4/8-bit 5,000 samples across datasets. {\DPQ} demonstrates enhanced performance and computational efficiency over TFMQ-DM and Q-Diff.
}
\label{tab:throughput_all}
\end{table*}

\begin{table}[t!]  
\centering
\centering\resizebox{\linewidth}{!}{
\begin{tabular}{@{}ccccccc@{}}
\midrule
\textbf{Base Method} & \textbf{Collection} & \textbf{P ($\uparrow$)} & \textbf{R ($\uparrow$)} & \textbf{F1 ($\uparrow$)} & \textbf{IS ($\uparrow$) / FID ($\downarrow$)} \\ 
\midrule
\multirow{2}{*}{\shortstack{ImageNet\\PTQ4DM}} 
& Deterministic & 0.5789 & 0.6014 & 0.5900 & 15.41/24.45 \\ 
& \cellcolor[HTML]{e6e6e6}{\bf Stochastic} & \cellcolor[HTML]{e6e6e6} \bf{0.6175} & \cellcolor[HTML]{e6e6e6} \bf{0.6201} & \cellcolor[HTML]{e6e6e6} \bf{0.6188} & \cellcolor[HTML]{e6e6e6} \bf{15.60/22.78} \\ 
\hdashline \vspace{-0.3cm}\\
\multirow{2}{*}{\shortstack{ImageNet\\Q-Diff}} 
& Deterministic & 0.6023 & 0.6189 & 0.6105 & 15.38/12.05 \\ 
& \cellcolor[HTML]{e6e6e6}{\bf Stochastic} & \cellcolor[HTML]{e6e6e6} \bf{0.6345} & \cellcolor[HTML]{e6e6e6} \bf{0.6372} & \cellcolor[HTML]{e6e6e6} \bf{0.6358} & \cellcolor[HTML]{e6e6e6} \bf{15.91/10.56} \\ 
\hdashline \vspace{-0.3cm}\\
\multirow{2}{*}{\shortstack{ImageNet\\TFMQ-DM}} 
& Deterministic & 0.6178 & 0.6285 & 0.6225 & 15.78/10.45 \\ 
& \cellcolor[HTML]{e6e6e6}{\bf Stochastic} & \cellcolor[HTML]{e6e6e6} \bf{0.6409} & \cellcolor[HTML]{e6e6e6} \bf{0.6444} & \cellcolor[HTML]{e6e6e6} \bf{0.6423} & \cellcolor[HTML]{e6e6e6} \bf{16.07/9.89} \\ 
\midrule
\end{tabular}}
\caption{
Classification accuracy (P: Precision, R: Recall, F1) and image generation quality (IS, FID) for different calibration data collection methods (Deterministic vs. Stochastic) across PTQ4DM, Q-Diff and TFMQ-DM with $4/8$-bit quantization on ImageNet DDIM.
}
\label{tb:comparison_der_or_stoc}
\end{table}

\subsection{Ablation Study}
\noindent \textbf{Impact of $\set{C}_\text{DC}$ Composition.} 
We assessed the effect of stochastic versus deterministic sampling for constructing the calibration dataset $\set{C}_\text{DC}$ in CAD phase. Stochastic sampling introduces noise variability, avoiding overfitting and improving generalization, as discussed in Sec.~\ref{sec:dataset}. 
The downstream task involved classification accuracy on ImageNet using $10$K generated samples evaluated with an Inception v3 model. 
As shown in Tab.~\ref{tb:comparison_der_or_stoc}, stochastic sampling outperforms deterministic sampling with up to $4.88\%$ higher F1 scores, demonstrating its efficacy in preventing biased generation.

\begin{figure}[t!]
\centering{\includegraphics[width=\linewidth]{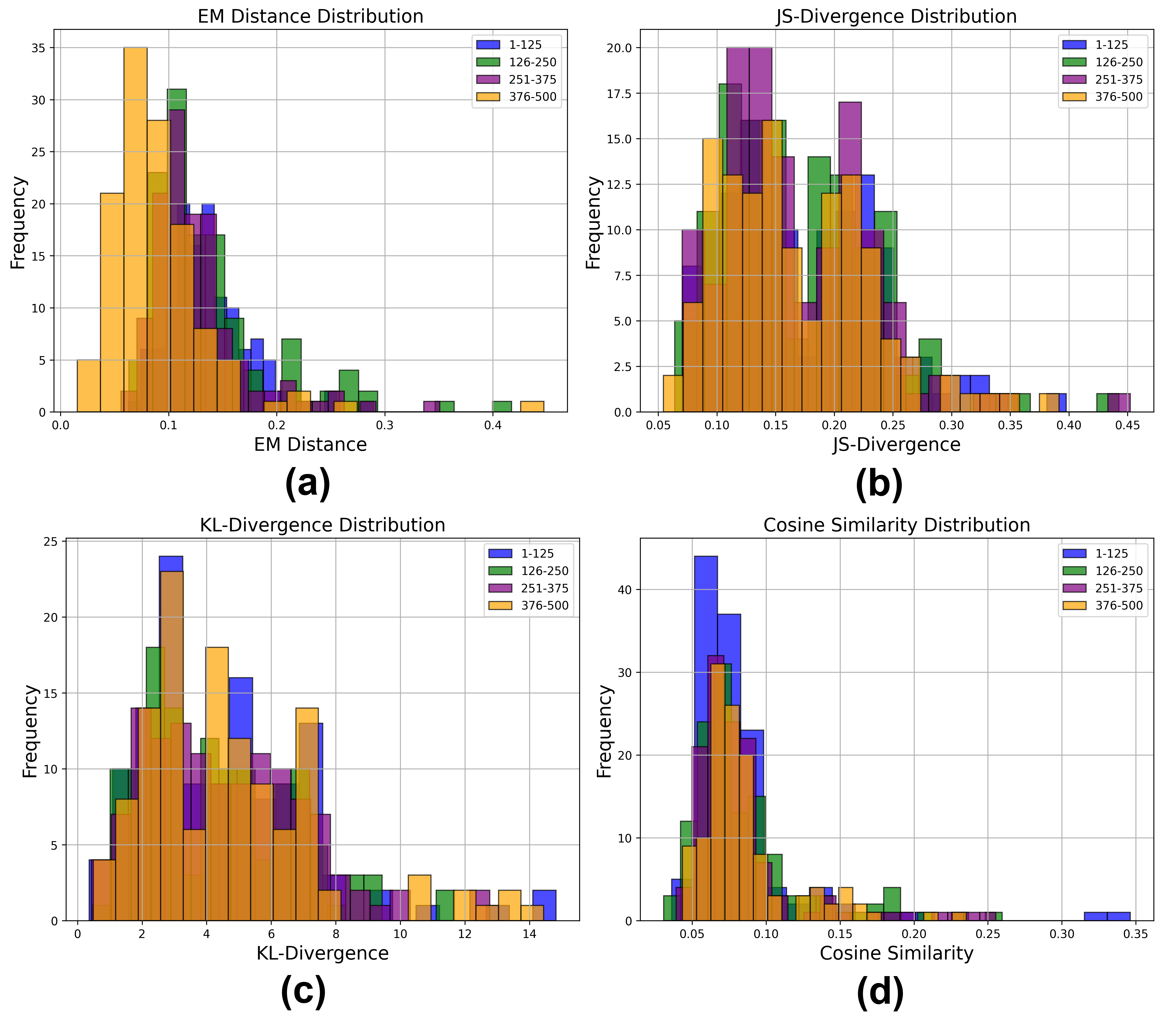}}
  \caption{Changes in $L_{\text{CD}}$ during the 500-step distillation process, plotted with respect to distance. Colors represent different distillation stages. Results are based on unconditional image generation with a 256×256 LSUN-Bedrooms LDM at 8/8-bit 100 sampling steps.
}
  \label{fig:compare_em_kl_jsd}
\end{figure}

\noindent \textbf{Distance Measures in $Loss_\text{CD}$.} 
We analyzed the performance of various distance measures for $Loss_\text{CD}$ in \eqref{eq:CAD}. 
Cosine similarity underperformed due to its inability to measure probabilistic distributions. In contrast, KL and JSD divergence metrics demonstrated slight improvements but were less effective in handling non-overlapping distributions. 
The Earth-Mover distance (EM) provided the most stable and meaningful measures of similarity, resulting in significant improvements in generative quality. 
As shown in Tab.~\ref{tab:cad_methods}, applying EM distance reduced FID and sFID by $6.69$ and $10.06$, respectively, when using TFMQ-DM as a base method for quantization. We further validate this observation by visualizing the evolution of each metric during distillation (Fig.~\ref{fig:compare_em_kl_jsd}). Cosine similarity failed to capture distributional changes due to its directional nature, while KL and JSD showed only modest sensitivity. EM distance, however, yielded clear convergence signals, supporting its effectiveness in aligning distributions and enhancing generation quality.



\begin{table}[t!]
    \centering
    \resizebox{0.95\linewidth}{!}{
    \begin{tabular}{@{}lcccc@{}}
        \toprule
        \textbf{CAD Methods} & \textbf{Base Method} & \textbf{Bits (W/A)} &\textbf{FID$\downarrow$} & \textbf{sFID$\downarrow$} \\
        \midrule
        Only $Loss_\text{KD}$& Full Prec.&32/32  &  3.07  & 7.15 \\
        \midrule
         \multirow{2}{*}{\shortstack{Only $Loss_\text{KD}$}} & Q-Diff &\multirow{2}{*}{\shortstack{8/8}}& 6.12 & 10.75  \\
         & TFMQ-DM & & 4.97 & 9.55 \\ \cmidrule(l){2-5}
        \multirow{2}{*}{\shortstack{+ $Loss_\text{CD}$ (Cos-Dist)}}   &  Q-Diff &\multirow{2}{*}{\shortstack{8/8}}  & 8.95 & 12.56  \\
         &  TFMQ-DM &  & 6.47 & 11.45  \\ \cmidrule(l){2-5}
         \multirow{2}{*}{\shortstack{+ $Loss_\text{CD}$ (KL-Div)}}        &  Q-Diff &\multirow{2}{*}{\shortstack{8/8}}  & 5.75 & 9.83  \\
         &  TFMQ-DM &  & 4.73 & 9.26 \\ \cmidrule(l){2-5}
         \multirow{2}{*}{\shortstack{+ $Loss_\text{CD}$ (JSD-Div)}}    &  Q-Diff &\multirow{2}{*}{\shortstack{8/8}}  & 5.18 & 9.49  \\
         &  TFMQ-DM  & & 4.59 & 9.04 \\ \cmidrule(l){2-5}
         \multirow{2}{*}{\shortstack{+ $Loss_\text{CD}$ (EM-Dist)}}  &  Q-Diff &\multirow{2}{*}{\shortstack{8/8}}   & \cellcolor[HTML]{e6e6e6} \textbf{4.36 \textcolor{blue}{(-1.76)}} &\cellcolor[HTML]{e6e6e6}  \textbf{7.88 \textcolor{blue}{(-2.87)}}  \\
         &  TFMQ-DM &  & \cellcolor[HTML]{e6e6e6} \textbf{3.10 \textcolor{blue}{(-1.87)}}& \cellcolor[HTML]{e6e6e6} \textbf{7.12 \textcolor{blue}{(-2.43)}}  \\ 
        \bottomrule
    \end{tabular}}
    \caption{
    Performance of CAD with different $Loss_\text{CD}$ settings using teacher model quantized by Q-Diff/TFMQ-DM on LSUN-Bedrooms LDM. \textcolor{blue}{Blue} indicates improvements relative to baseline. 
    }
    \label{tab:cad_methods}
\end{table}

\vspace{-0.15cm}
\section{Discussion and future work}

\section{Conclusion}

In this work, we introduced {\DPQ}, a hybrid framework combining progressive quantization and calibration-assisted distillation to compress diffusion models effectively. By leveraging dual calibration datasets, PQ and CAD, our method minimizes error accumulation and balances computational efficiency (halving inference time) with generative performance. 
Experimental results demonstrate that {\DPQ} achieves significant reductions in memory and sampling steps while maintaining competitive generative quality across diverse datasets and architectures. 
Our approach highlights the importance of tailored calibration and sequential optimization for compressing diffusion models, paving the way for future research in efficient activation quantization.




\bibliographystyle{IEEEbib}
\bibliography{ref}

\begin{IEEEbiography}[{\includegraphics[width=1in,height=1.25in,clip,keepaspectratio]{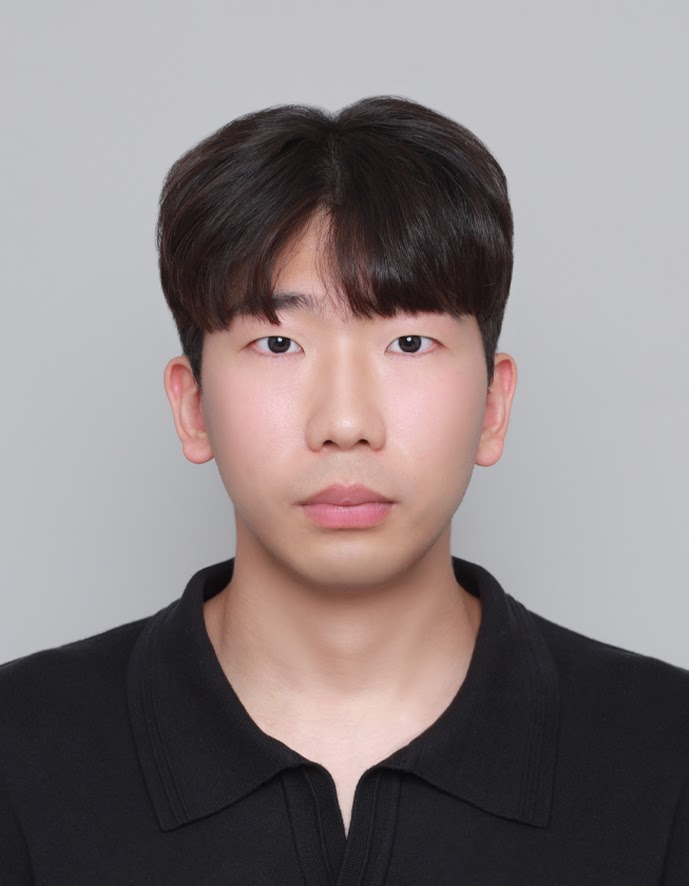}}]{Beomseok Ko}
(Student Member, IEEE) received the B.S. degree in Information and Communication Engineering from Dongguk University, Seoul, South Korea, in 2023. He is currently pursuing the M.S. degree in Computer Science and Artificial Intelligence at Dongguk University. His research interests include model compression and the application of diffusion models.
\end{IEEEbiography}

\begin{IEEEbiography}[{\includegraphics[width=1in,height=1.25in,clip,keepaspectratio]{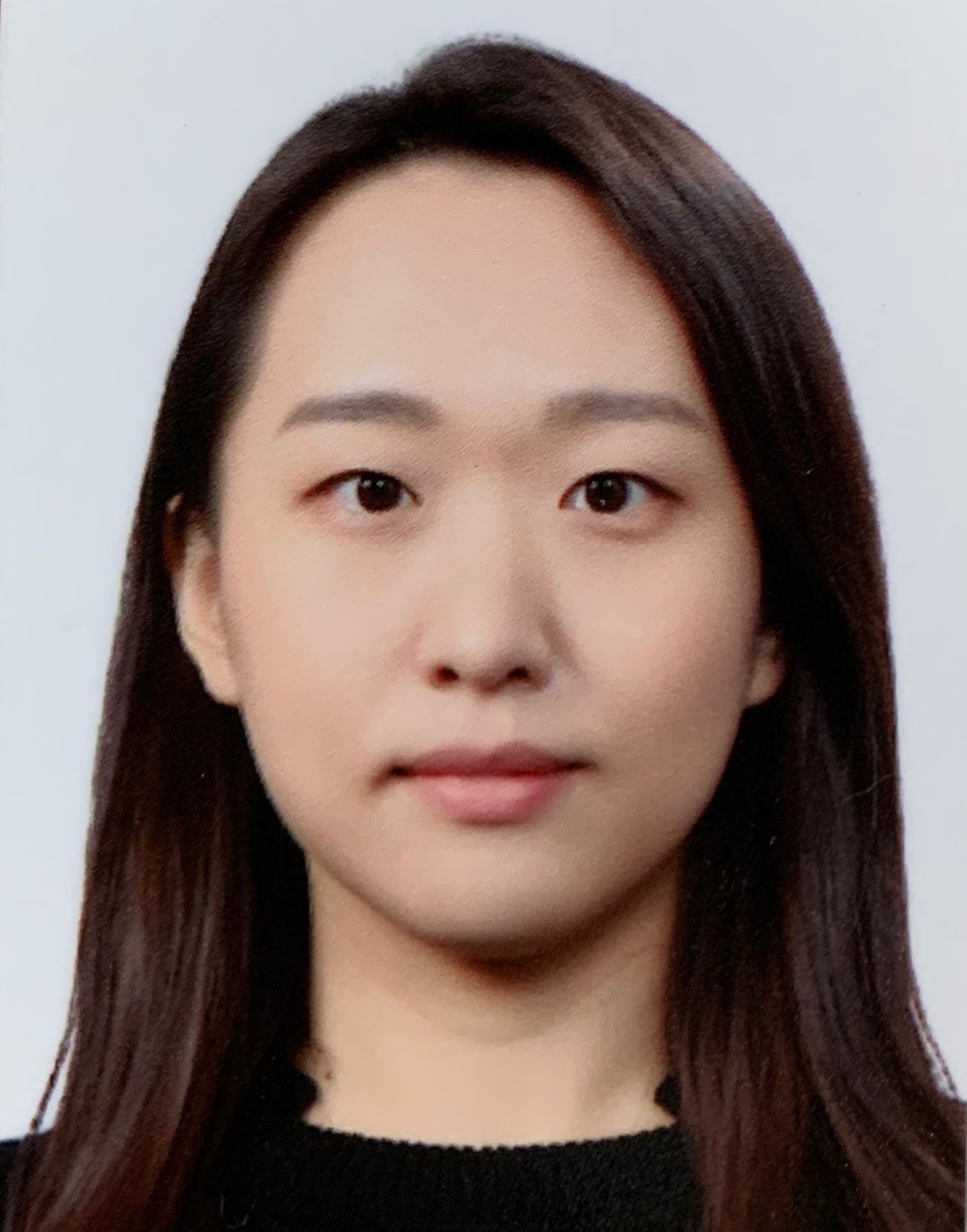}}]{Hyeryung Jang}
(Member, IEEE) received the B.S., M.S., and Ph.D. degrees from the School of Electrical Engineering, Korea Advanced Institute of Science and Technology (KAIST), Daejeon, South Korea, in 2010, 2012, and 2017, respectively. From 2018 to 2021, she was a Research
Associate with the Department of Informatics, King’s College London, London, U.K. She is currently an Assistant Professor with the Division of Computer Science and Artificial Intelligence, Dongguk University, Seoul, South Korea. Her research interests include learning and inference of probabilistic graphical models, machine learning on networked systems, wireless communications, and neuromorphic computing.
\end{IEEEbiography}

\end{CJK}
\end{document}